\documentclass{ifac/ifacconf}

\usepackage{graphicx}      
\usepackage{natbib}        
\usepackage{url}
\usepackage{comment}
\usepackage{xcolor}
\usepackage{subcaption}
\usepackage{amsmath}
\usepackage{lipsum}
\usepackage{mathtools}

\begin{document}

\newcommand{\todo}[1]{\textcolor{red}{#1}}

\begin{frontmatter}

\title{Explainable Anomaly
Detection for Electric Vehicles Charging Stations\thanksref{footnoteinfo}} 

\thanks[footnoteinfo]{This work was partially carried out within the Italian National Center for Sustainable Mobility (MOST) and received funding from NextGenerationEU (Italian NRRP – CN00000023 - D.D. 1033 17/06/2022 - CUP C93C22002750006).}

\author[First]{Matteo Cederle,} 
\author[First]{Andrea Mazzucco,} 
\author[Second]{Andrea Demartini,}
\author[Second]{Eugenio Mazza,}
\author[Second]{Eugenia Suriani,}
\author[Second]{Federico Vitti,}
\author[First]{Gian Antonio Susto}

\address[First]{University of Padova, 
   Italy (e-mail: matteo.cederle@phd.unipd.it, andrea.mazzucco.2@studenti.unipd.it, gianantonio.susto@unipd.it).}
\address[Second]{A2A S.p.A., 
   Italy (e-mail: andrea.demartini@a2a.it, eugenio.mazza@a2a.it, eugenia.suriani@a2a.it, federico.vitti@a2a.it).}

\begin{abstract}                
Electric vehicles (EV) charging stations are one of the critical infrastructures needed to support the transition to renewable-energy-based mobility, but ensuring their reliability and efficiency requires effective anomaly detection to identify irregularities in charging behavior. However, in such a productive scenario, it is also crucial to determine the underlying cause behind the detected anomalies. To achieve this goal, this study investigates unsupervised anomaly detection techniques for EV charging infrastructure, integrating eXplainable Artificial Intelligence techniques to enhance interpretability and uncover root causes of anomalies.

Using real-world sensors and charging session data, this work applies Isolation Forest to detect anomalies and employs the Depth-based Isolation Forest Feature Importance (DIFFI) method to identify the most important features contributing to such anomalies. The efficacy of the proposed approach is evaluated in a real industrial case.
\end{abstract}


\end{frontmatter}

\section{Introduction}
\label{sec:intro}

According to a recent report published by the association Motus-E, at the end of 2024 the number of installed electric vehicles (EV) charging stations in Italy amounted to 64.391, an increment of 27\% with respect to 2023, consistent with the important growth trend registered in the last years (\cite{motus2020infrastrutture}).
This rapidly evolving landscape necessitates an efficient monitoring and management of the EV charging infrastructure, essential for optimizing station performance and ensuring reliable service availability. Detecting irregularities in charging station operations, which translate to hardware failures, charging process issues, power fluctuations, or connectivity disruptions, is critical for minimizing downtime, reducing maintenance costs, and improving user experience.

In this context, anomaly detection (AD) plays a fundamental role in identifying unexpected faults, failures, or inefficiencies (\cite{chandola2009anomaly}). Unsupervised AD methods (\cite{goldstein2016comparative}) are particularly valuable in this context since they do not rely on labeled data, which can be difficult and time-consuming to obtain. Labeling typically requires human expertise and manual effort, making it impractical in large-scale deployments. By leveraging unsupervised approaches, it becomes possible to detect anomalies without the need for predefined fault classifications.
Moreover, in the EV charging infrastructure scenario it is crucial to understand the underlying causes behind the detected anomalies. This can be addressed through eXplainable Artificial Intelligence (XAI), which aims to provide insights into the reasoning behind machine learning (ML) based decisions. While some ML models offer inherent interpretability, high-performance methods often lack transparency in their decision-making processes. This opacity can reduce trust in ML-driven systems, posing a challenge for their integration into critical Decision Support Systems used for managing EV charging networks.
This is the case for current state-of-the-art AD techniques for EV charging stations, which employ advanced deep neural network architectures, such as ResNet Autoencoder, Bidirectional LSTM and ensemble of CNN-LSTM modules (\cite{mavikumbure2023physical, hussain2024anomaly, dixit2022anomaly}). In parallel to these works, \cite{cumplido2022collaborative} employed standard AD algorithms like Decision Trees and Random Forests to address a collaborative anomaly detection system for charging stations, which however also lacked explainability principles both for users and domain experts.

In light of that, in this work we propose to address the aforementioned challenges by applying, for the first time in a real-world EV charging stations scenario, an explainability-driven unsupervised anomaly detection algorithm, i.e. Depth-Based Isolation Forest Feature Importance (DIFFI, \cite{carletti2023interpretable}). This algorithm has been chosen for its proved effectiveness and reduced computational costs with respect to traditional explainability methods such as SHAP (\cite{lundberg2017unified}), which make it appealing for a real-world time sensitive application like the one that we are treating in this work.

The key contributions of this paper can be summarized as follows:

\begin{enumerate}
    \item We develop a well-structured tabular representation of a real-world EV charging station dataset, refining raw data through preprocessing, to ensure compatibility with anomaly detection methods like Isolation Forest (IF, \cite{liu2008isolation}), enabling more effective identification of irregular charging behaviors.
    \item To maximize efficiency and allow for real-time implementations, we use IF to detect anomalies and DIFFI to produce explanations.
    \item The proposed approach is evaluated in a real-world case study, by exploiting the infrastructure of a2a, an Italian company leader in the EV charging stations field.
\end{enumerate}

The remainder of this manuscript is structured as follows: Section \ref{sec:xai} provides some background about unsupervised and explainable anomaly detection, focusing in particular on the DIFFI algorithm, while Section \ref{sec:caseexp} presents in detail the specific case study of this work, along with the dataset used for our experiments and the necessary preprocessing steps. We also showcase the results achieved by applying the DIFFI algorithm to the EV charging station network and provide useful insights both from a global and a local pespective. Finally, Section \ref{sec:concl} concludes our study by providing additional comments and sketching some directions for future improvements.

\section{Explainable Anomaly Detection}
\label{sec:xai}

In this section, we first summarize the key concepts at the core
of the Isolation Forest algorithm and introduce the necessary notation. Then, we discuss the rationale behind the DIFFI method and analyze it in depth. We also describe the key differences between Global DIFFI and Local-DIFFI, a variant useful for the interpretation of individual
predictions.

\subsection{Isolation Forest}

The Isolation Forest algorithm is an  unsupervised anomaly detection method that leverages recursive partitioning to derive an \textit{anomaly score} for each data point, which quantifies its degree of outlierness within the dataset. IF aims to define isolated regions within the data domain where only individual points reside. The core hypothesis underpinning the IF algorithm is that isolating outliers necessitates fewer iterations compared to the extensive recursive partitions required for inliers.

The architecture of IF consists of an ensemble of $T$ \textit{Isolation Trees (ITs)} $\lbrace t_1, \dots, t_T  \rbrace$, each structured as a random tree. They are constructed by randomly selecting, at each internal node $v$, both a splitting feature and its threshold. For a given dataset $\mathcal{D} = \lbrace \mathbf{x}_1, \dots, \mathbf{x}_n \rbrace$ of $d$-dimensional points, each IT operates on a distinct bootstrap sample $\mathcal{D}_t\subset\mathcal{D}$. The isolation process continues recursively until a predefined maximum depth limit is reached or further splitting yields no additional information.

The anomaly score for any point $\mathbf{x}_i$ is then computed through a function proportional to the average path length of $\mathbf{x}_i$ across all trees in the ensemble, ensuring that the anomaly scores reflect both the isolation efficiency and the data structure. In its concluding phase, the IF algorithm employs a thresholding operation on the computed anomaly scores to classify points as either anomalous or normal. This partitioning of the dataset results in two subsets: the predicted inliers $\mathcal{P}_I = \lbrace \mathbf{x}_i \in \mathcal{D} \, | \, \hat{y}_i = 0  \rbrace$ and the predicted outliers $\mathcal{P}_O = \lbrace \mathbf{x}_i \in \mathcal{D} \, | \, \hat{y}_i = 1  \rbrace$, with binary labels $\hat{y}_i \in \lbrace 0,1 \rbrace$ indicating the respective classifications.

\subsection{DIFFI algorithm}
\label{subsec:diffi}

DIFFI is an unsupervised model-specific approach developed to interpret the IF algorithm, utilizing its inherent tree structure. It identifies crucial features for determining whether a data point is an outlier or not by asserting that relevant features should promptly isolate anomalies and create significant imbalances when isolating anomalous points, whereas inliers exhibit the converse behavior.

To illustrate DIFFI's functionality, we present several components, starting from the \textit{induced imbalance coefficients (IIC)} $\lambda$: for an internal node $k$ within an isolation tree, let $n(k)$ represent the total number of divided points, and $n_l(k)$ and $n_r(k)$ denote the counts in its left and right child nodes, respectively. The IIC for a node $v$ is then given by:

\begin{equation}
    \lambda(v)=\begin{cases}
    0 & \text{if } n_l(k)=0 \text{ or } n_r(k)=0 \\
    \tilde{\lambda}(k) & \text{otherwise}
\end{cases}
\end{equation}

where \\ {\small$\tilde{\lambda}(k)=g(\frac{\max(n_l(k),n_r(k))}{n(k)})$, and $g(a)=\frac{a-\lambda_{min}(n)}{2(\lambda_{max}(n)-\lambda_{min}(n))}+0.5$.} \\ Here, $\lambda_{min}$ and $\lambda_{max}$ denote the minimum and maximum scores possible given the number of data points $n(k)$.

Furthermore, we introduce the \textit{cumulative feature importances I} as vectors with dimensionality equal to the feature count $d$, where each component denotes the importance of the corresponding feature. We distinguish between $I_I$ for predicted inliers and $I_O$ for outliers. Considering $I_I$, initialization starts with $I_I=0_d$. Given the subset $\mathcal{P}_{I,t}$ of predicted inliers within tree $t$, DIFFI iterates over the internal nodes along each inlier's path $Path(x_I,t)$. If a node $v$ uses feature $f_j$ for splitting, then the $j$-th component of $I_I$ is updated by adding the following quantity:

\begin{equation}
    \Delta=\cfrac{1}{h_t(x_I)}\lambda_I(v),
\end{equation}
where $h_t(x_I)$ denotes the depth of $x_I$ in tree $t$.

An analogous procedure applies for $x_O\in\mathcal{P}_{O,t}$ and $I_O$. Intuitively, this means that features that isolate points
sooner are considered to be more useful.

Finally, the \textit{features counter V} addresses potential biases from uneven random feature sampling affecting cumulative feature importance calculations. Each time a point passes through a node, its corresponding splitting feature counter increments by one. Similar to the cumulative feature importance vectors, two distinct counters, $V_I$ for inliers and $V_O$ for outliers, are maintained.

In conclusion, DIFFI determines Global Feature Importance by assessing the weighted ratio of outlier-to-inlier cumulative feature importances:

\begin{equation}
    GFI=\cfrac{I_O/V_O}{I_I/V_I}.
\end{equation}

For the interpretation of individual predictions produced by the IF, \cite{carletti2023interpretable} proposed also a local version of DIFFI, which enables the interpretation of single data points in online settings when the model has already been deployed, and also helps to enhance trust in the model, as the user can check whether the algorithm tends to make mistakes on those kinds of inputs where humans also make mistakes. We refer to the original paper for a technical explanation of Local-DIFFI.

\section{Dataset and Experiments}
\label{sec:caseexp}

As already mentioned in Section \ref{sec:intro}, for this study we have exploited a real-world dataset, produced by the Italian company a2a. Specifically, we have considered data coming from eight EV charging stations in the time span between January 2022 and June 2024, for a total of $22986$ full recharges\footnote{The actual number of recharges in the selected time span is higher, but we removed from the analysis recharges with missing or incomplete information.}.

The available data for each recharge primarily included tabular metrics, such as the \textit{mean power delivered} $P_{mean}$, the \textit{mean energy delivered} $E_{mean}$, the \textit{mean temperature during the recharge} $T_{mean}$, the \textit{total CO$_2$ emissions} in kilograms, and the \textit{duration of the recharge}. 
These parameters provide a summary view of the main features of a charging cycle and enable comparisons across different cycles at an aggregated level.
Additionally, discretized signals representing instantaneous power and temperature were also recorded for each session, providing detailed and dynamic information about the recharge. To leverage this information in algorithms like Isolation Forest, which require tabular input, a pre-processing step was performed to extract meaningful features. This transformation not only made the data compatible with IF but also provided a more concise representation of the recharge behavior.

The full set of features used for this work is reported in Table \ref{tb:feat}.

\begin{table}[h]
\begin{center}
\caption{Available features for each recharge}\label{tb:feat}
\begin{tabular}{l|l}
\hline
mean power $P_{mean}$ & mean energy $E_{mean}$\\
mean temperature $T_{mean}$ & kilograms of emitted CO$_2$\\
time duration & correlation P-T $Corr_{P-T}$\\
power standard dev. $P_{std}$ & temperature standard dev. $T_{std}$ \\
power maximum $P_{MAX}$ & temperature maximum $T_{MAX}$ \\
power minimum $P_{min}$ & temperature minimum $T_{min}$ \\
power abs. diff. $P_{absdiff}$ & temperature abs. diff $T_{absdiff}$ \\
power skewer $P_{skewer}$ & temperature skewer $T_{skewer}$\\
power kurtosis $P_{kurt}$ & temperature kurtosis $T_{kurt}$\\
power num. peaks $P_{npeaks}$ & temperature num. peaks $T_{npeaks}$\\
power corr. lag $P_{corrlag}$ & temperature corr. lag $T_{corrlag}$\\
\hline
\end{tabular}
\end{center}
\end{table}

\subsection{Global feature ranking and selection}
\label{subsec:global}

\begin{figure}[ht]
\begin{center}
\includegraphics[width=8.4cm]{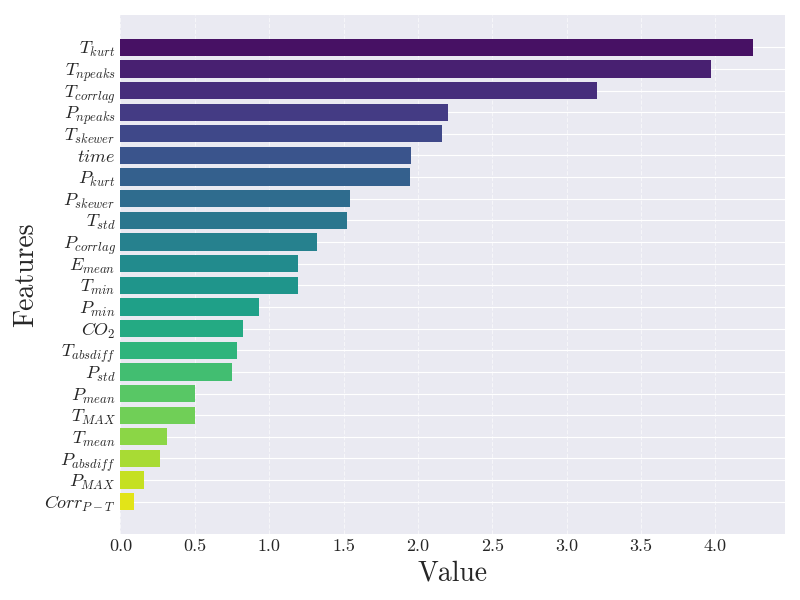}
\caption{DIFFI Global Feature Importance scores.} 
\label{fig:global}
\end{center}
\end{figure}

\begin{figure*}[t]
    \centering
    \begin{subfigure}[b]{0.48\textwidth}
        \centering
        \includegraphics[width=\textwidth]{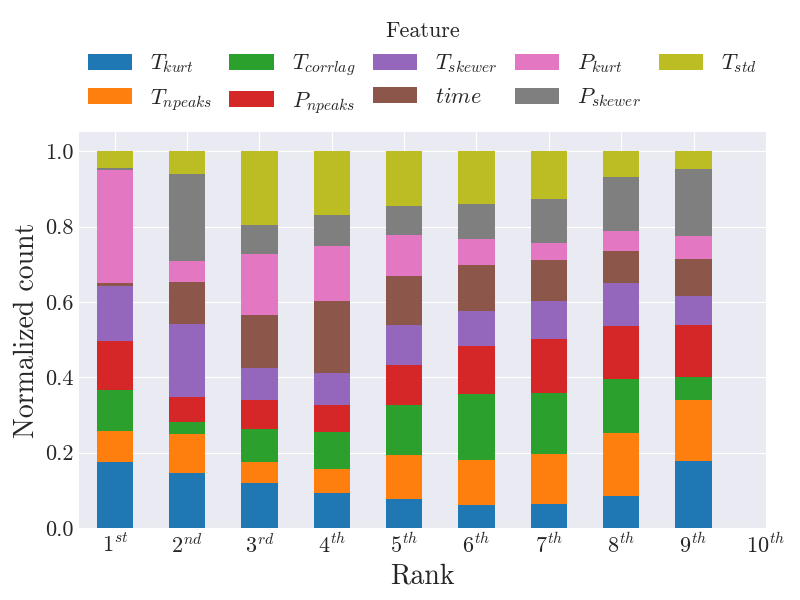}
        \caption{Local-DIFFI}
        \label{fig:first}
    \end{subfigure}
    \hfill
    \begin{subfigure}[b]{0.48\textwidth}
        \centering
        \includegraphics[width=\textwidth]{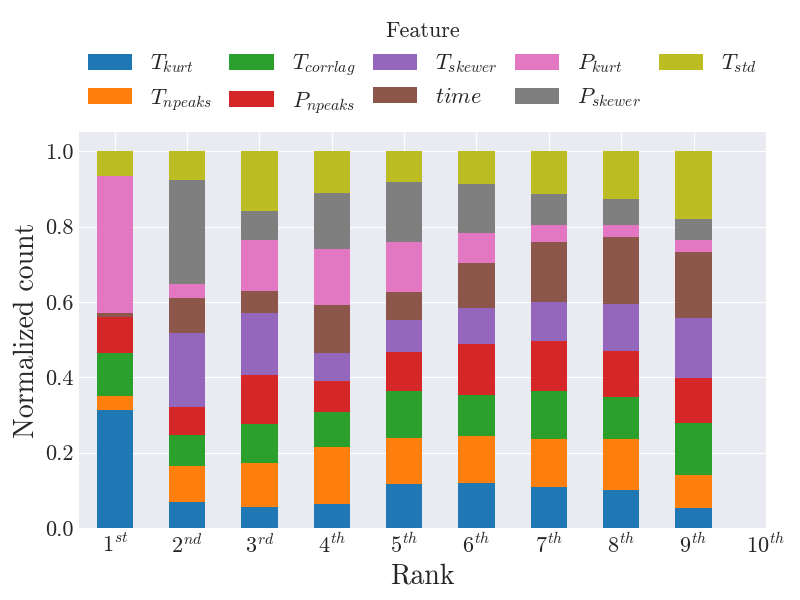}
        \caption{SHAP}
        \label{fig:second}
    \end{subfigure}
    
    \caption{Feature rankings based on Local-DIFFI scores (a) and SHAP scores (b)}
    \label{fig:comparison}
\end{figure*}

As an initial analysis, we employed the global DIFFI algorithm to evaluate feature importance, as illustrated in Figure \ref{fig:global}. Looking at the plot, it is evident that the top three features, i.e. \textit{temperature kurtosis}, \textit{temperature num. peaks}, and \textit{temperature correlation lag} played the most significant role in predicting the outputs of the global DIFFI.
We remark that the aim of this analysis is to understand the influence of each feature on a global scale and, if necessary, perform unsupervised feature selection by eliminating the least significant features. This approach can enhance model interpretability and reduce complexity. Specifically, for the proceeding of our analysis, we decided to maintain only the nine most important features, following the ranking displayed in Figure \ref{fig:global}. 

\subsection{Local interpretation of individual predictions}
To interpret the individual predictions made by the IF model, we employed the Local-DIFFI method, outlined briefly in Section \ref{subsec:diffi}. This was combined with the reduced set of features, which were selected through the unsupervised feature selection process described in Section \ref{subsec:global}.
We remark that this setup is extremely useful in a real scenario like the one we are considering: once the Isolation Forest model is trained, the user aims to deploy it in an online setting, where it can provide predictions along with the corresponding local feature importance scores for each individual data point being processed.

Figure \ref{fig:first} shows the results of the Local-DIFFI method, applied to the anomalous points detected by the IF algorithm. Here, features are color-coded, with columns representing the ranks, and the height of the bar associated with each feature representing the fraction of predicted anomalies for which the feature has a specific rank.

We remark that the real-world dataset considered for this work does not provide labels for anomalies, making the validation of anomaly detection algorithms extremely challenging. However, our findings were discussed with domain experts, which found the interpretability results reasonable and in line with available domain knowledge in the majority of the cases.
A further validation of the effectiveness of our results is provided by comparing Local-DIFFI and SHAP, as shown in Figure \ref{fig:comparison}. SHAP is widely recognized as a prominent explainability method for anomaly detection algorithms and is frequently used in the literature to validate other approaches, especially when ground truth labels are unavailable. As illustrated in the plots, both DIFFI and SHAP identified similar influential features. However, DIFFI offers a notable advantage by delivering results at a considerably lower computational cost (\cite{carletti2023interpretable}).
This makes DIFFI particularly well-suited for a large-scale, time-sensitive anomaly detection task, such as the one encountered in the context of electric vehicles charging stations.

\section{Conclusions}
\label{sec:concl}

This study explored explainable anomaly detection techniques for electric vehicles charging stations, focusing on identifying irregular charging behaviors and interpreting their root causes. Using real-world data, we applied the Isolation Forest algorithm to detect anomalies and employed the Depth-based Isolation Forest Feature Importance method to determine the most influential factors contributing to these deviations. Our findings highlight key features associated with anomalous charging sessions, offering insights into potential faults, inefficiencies, or security concerns.

Beyond anomaly detection, this work could inspire predictive maintenance strategies in the EV charging stations field, enabling proactive interventions before failures occur (\cite{lorenti2023predictive}). Such an approach would enhance charging stations reliability, reduce operational costs, and contribute to a more efficient EV ecosystem.


\bibliography{main}

\end{document}